\documentclass[10pt,twocolumn,letterpaper]{article}

\usepackage{cvpr}              

\usepackage{graphicx}  
\usepackage{booktabs} 
\usepackage{tabularx}  
\usepackage{makecell} 
\usepackage{array}
\usepackage{threeparttable}
\usepackage{multirow}
\usepackage{xcolor}
\usepackage{float}
\usepackage{caption}  
\usepackage{placeins}

\definecolor{cvprblue}{rgb}{0.21,0.49,0.74}
\usepackage[pagebackref,breaklinks,colorlinks,allcolors=cvprblue]{hyperref}

\def\paperID{*****} 
\def\confName{CVPR}
\def\confYear{2025}

\title{MedErr-CT: A Visual Question Answering Benchmark for Identifying and Correcting Errors in CT Reports}

\author{Sunggu Kyung, Hyungbin Park, Jinyoung Seo, Jimin Sung, Jihyun Kim, Dongyeong Kim, \\
Wooyoung Jo, Yoojin Nam, Sangah Park, Taehee Kwon, Sang Min Lee, Namkug Kim\\
\\
Department of Biomedical Engineering, University of Ulsan College of Medicine\\
Seoul, Republic of Korea\\
{\tt\small babbu3682@gmail.com, namkugkim@gmail.com}
}

\begin{document}
\maketitle
\begin{abstract}
Computed Tomography (CT) plays a crucial role in clinical diagnosis, but the growing demand for CT examinations has raised concerns about diagnostic errors. While Multimodal Large Language Models (MLLMs) demonstrate promising comprehension of medical knowledge, their tendency to produce inaccurate information highlights the need for rigorous validation. However, existing medical visual question answering (VQA) benchmarks primarily focus on simple visual recognition tasks, lacking clinical relevance and failing to assess expert-level knowledge. We introduce MedErr-CT, a novel benchmark for evaluating medical MLLMs' ability to identify and correct errors in CT reports through a VQA framework. The benchmark includes six error categories—four vision-centric errors (Omission, Insertion, Direction, Size) and two lexical error types (Unit, Typo)—and is organized into three task levels: classification, detection, and correction. Using this benchmark, we quantitatively assess the performance of state-of-the-art 3D medical MLLMs, revealing substantial variation in their capabilities across different error types. Our benchmark contributes to the development of more reliable and clinically applicable MLLMs, ultimately helping reduce diagnostic errors and improve accuracy in clinical practice. The code and datasets are available at : \href{https://github.com/babbu3682/MedErr-CT}{https://github.com/babbu3682/MedErr-CT} 
\end{abstract}    
\section{Introduction}
\label{sec:intro}
\begin{figure*}[t]
  \centering
  \includegraphics[width=\textwidth]{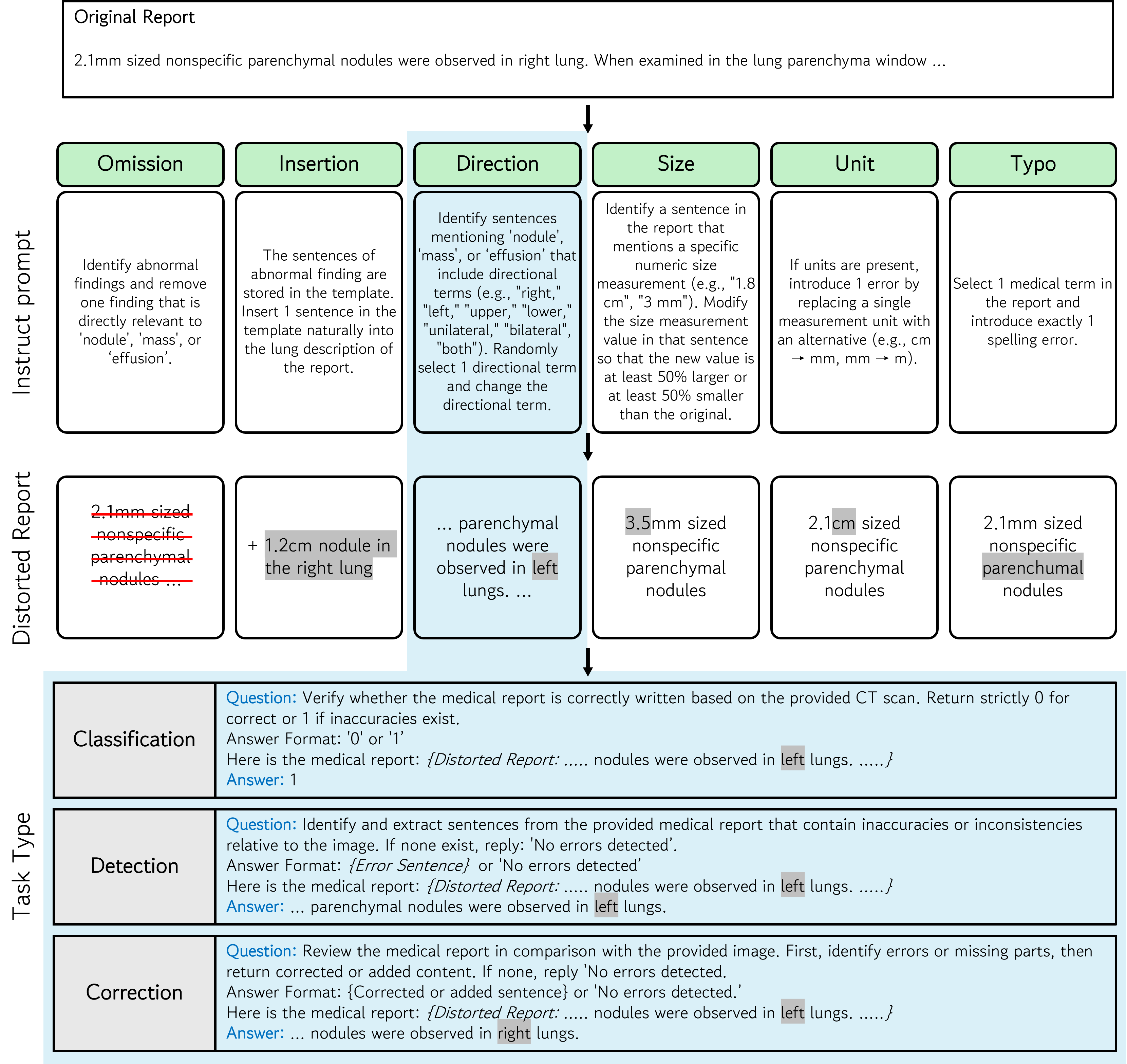}
  \caption{The data generation pipeline of the MedErr-CT Benchmark. We automated data generation utilizing the DSPy framework with LLaMA-3.3-70B. Through prompts corresponding to six error types, distorted reports were generated for each error category, which were then used to compose QA sets. The benchmark consists of three difficulty levels, with each stage evaluating the model's medical reasoning capabilities in depth: (1) a classification task to determine the presence of errors in CT-based reports, (2) a detection task to identify sentences containing errors, and (3) a correction task to rectify erroneous statements.}
  \label{fig:one}
\end{figure*}
Computed Tomography (CT) imaging plays a pivotal role in clinical diagnosis by offering detailed visualization of internal structures crucial for accurate disease identification and treatment planning \cite{muller2002computed}. However, the rising demand for CT examinations has substantially intensified the workload for radiologists \cite{brenner2010should}. This increased workload has raised concerns about diagnostic fatigue and errors. Lee \textit{et al.} \cite{lee2013cognitive} reported a retrospective discrepancy rate as high as 30\%, suggesting that a considerable portion of abnormalities were initially overlooked or inaccurately described.

Recent advancements in Multimodal Large Language Models (MLLMs) have garnered significant popularity in the medical field \cite{sloan2024automated}. General MLLMs have achieved faster processing time and comparable performance to radiology experts in differential diagnosis, demonstrating strong potential as supportive tools in diagnostic decision-making \cite{Suh2024}. In the field of medical MLLMs, while 2D medical MLLMs have driven early progress, research on 3D medical MLLMs is rapidly advancing despite the inherent challenges posed by volumetric imaging modalities. These models are transforming how medical professionals interpret and document imaging findings. However, one of the most critical issues in integrating MLLMs into clinical practice is their tendency to generate misleading or inaccurate information \cite{tang2023evaluating}. Rigorous validation processes are essential to minimize these risks and safely utilize MLLMs for medical content generation \cite{karabacak2023embracing}.

Although several existing benchmarks \cite{lau2018dataset, he2020pathvqa, liu2021slake, zhang2023pmc, hu2024omnimedvqa, ye2024gmai} have been proposed to evaluate the medical knowledge of current MLLMs, they primarily focus on simple visual recognition, lacking relevance to real-world clinical scenarios \cite{zuo2025medxpertqa}. In particular, the situation is even more severe for 3D medical benchmarks. Consequently, they fall short in thoroughly evaluating medical expert-level knowledge and complex reasoning skills required for diagnostic decision-making. Such simplified evaluation protocols can lead developers to optimize models on basic question-answering format \cite{fu2024mmesurvey}, hindering their ability to solve real-world medical instruction-following tasks.

Clinical errors remain a prevalent and critical issue in healthcare. In particular, frequent radiologist errors types include missing abnormality, incorrect lesion localization \cite{Kimfoolme2014}.
Moreover, the risk of malpractice litigation, excessive workload, urgent medical scenarios, and a strong sense of professional duty collectively contribute to significant psychological pressure on clinicians regarding their mistakes \cite{patel2011recovery}.
A recent survey study from three US health care organizations showed that 1 in 5 patients who read a clinical note reported finding a mistake and 40\% perceived the mistake as serious \cite{BenAbachaMediqa2024}.
In this context, there is a pressing need for AI that actively helps them identify and reflect on possible mistakes, not simply an AI that can write medical reports well enough to replace clinicians.
Consequently, medical MLLMs must embody these capabilities and be evaluated accordingly.

Patel \textit{et al.} \cite{patel2011recovery} show clear differences in the ability of healthcare professionals to detect and recover from errors depending on their level of expertise. This suggests a strong association between clinical proficiency and the capacity to manage medical errors effectively.
Gertz \textit{et al.} \cite{gertz2024potential} reported that utilizing GPT-4 for medical reading error correction in real clinical environments resulted in improved workflow efficiency and effective cost reduction.
Abacha \textit{et al.} \cite{BenAbachaMediqa2024} demonstrated that automatic detection and correction of medical errors can improve the accuracy and consistency of clinical documentation, contributing to improved patient care and health outcomes.

Inspired by these prior findings on the correlation between clinical expertise and diagnostic error recovery, and the clinically practical  benefits of automation in medical error detection/correction, the present study proposes a novel benchmark for medical error. We evaluate the medical capabilities of open-source MLLMs through their ability to handle errors in radiology reports, adopting a visual question answering (VQA) format as the evaluation framework. Furthermore, we emphasize the evaluation of the important yet underexplored capability of automatic error detection and correction in radiology AI.

The benchmark includes a total of six error types—four vision-centric and two lexical error categories—which are further stratified into three levels of difficulty to enable a comprehensive and nuanced assessment.

Our contributions are as follows:
\begin{itemize}
   \item We developed a VQA dataset of six error categories including vision-centric errors based on CT imagings.

    \item We provide a multi-level evaluation framework that assesses model performance across three stages: error classification, detection, and correction.
    
   \item We perform a comprehensive evaluation of state-of-the-art 3D medical MLLMs through both quantitative and qualitative analyses.

    \item We evaluate the performance of clinically practical error VQA among state-of-the-art 3D medical MLLMs

\end{itemize}
\section{Related works}
\label{sec:formatting}

\subsection{Medical 3D MLLMs}

Multimodal large language models (MLLMs) have demonstrated remarkable performance improvements by utilizing large language models (LLMs) as decoders and applying instruction-following datasets \cite{chen2024next}. While the development of 3D medical MLLMs has been relatively limited due to the scarcity of suitable data \cite{chen20243d}, various open-source 3D medical MLLMs have been released recently \cite{wu2023towards,bai2024m3d,hamamci2024developing}. 

RadFM \cite{wu2023towards} is a pioneering radiology foundation model that combines 3D vision transformers (ViT) with Perceiver architecture to process both 2D and 3D scans across diverse imaging modalities. It employs MedLLaMA-13B as its language model. M3D \cite{bai2024m3d} utilized a 3D ViT encoder coupled with a spatial pooling perceiver to efficiently compress high-dimensional tokens while enabling rapid segmentation processing. It supports both LLaMA and Phi as its language backbones.
CT-CHAT \cite{hamamci2024developing} is a 3D multimodal AI assistant, adopting a CT-CLIP-based vision encoder with attention-pooling layers and multi-layer perceptron (MLP) layers, using LLaMA, Mistral, and Vicuna as language models.
Med3DVLM \cite{xin_med3dvlm} is a 3D MLLM that incorporates a DCFormer-based vision encoder, an MLP-Mixer connector, and the Qwen 2.5-7B-Instruct language model.
MedM-VL \cite{shi_medmvl} is a model that handles both 2D and 3D images. For 3D images, it adopts a SigLIP-based 2D vision encoder, a cross-attention connector module, and the Qwen 2.5-3B-Instruct language model.
\subsection{Medical VQA Dataset}

Visual Question Answering (VQA) provides an effective framework for evaluating the comprehensive understanding capabilities of MLLMs through question-and-answer paradigms.
Several benchmarks have been developed to evaluate VQA, including VQA-RAD, PMC-VQA , and OmniMedVQA , which are primarily based on 2D medical images. Although recent works such as RP3D-VQA and M3D-VQA have extended VQA tasks to 3D imaging, such datasets remain scarce. Moreover, these benchmarks are limited in the diversity of question types, focusing mainly on modality classification, anatomical localization, or disease identification—thus falling short in evaluating deeper comprehension of medical knowledge and reasoning skills essential for real-world clinical applications.

\subsection{Medical Error-centered Studies}
Several studies have introduced medical error-centered datasets to evaluate the error detection and correction capabilities of AI models in the medical domain.

MEDEC \cite{abacha2024medec} proposed five types of commonly observed errors and constructed a benchmark dataset by introducing errors into clinical notes. They evaluated the ability of state-of-the-art LLMs to detect errors, identify inaccurate statements, and correct them in medical documents. Additionally, Gertz \textit{et al.} evaluated the capability of GPT-4 to detect and correct reading errors \cite{gertz2024potential}. However, these studies rely solely on text input without integrating medical imaging data.

To address these limitations, ReXErr \cite{rao_rexerr_2024} generated error datasets from MIMIC-CXR \cite{johnson2019mimic} chest X-ray images and corresponding reports. They analyzed error patterns that are frequently observed in both human-written and AI-generated reports and utilized these insights to construct error reports. Unlike previous studies, ReXErr is explicitly designed to facilitate MLLM-based error detection and correction by providing paired datasets that include both radiological images and error reports. Nevertheless, ReXErr has several notable limitations. It is limited to 2D chest X-ray data and lacks scalability for 3D imaging modalities such as CT and MRI. Furthermore, the authors did not conduct empirical evaluations of MLLMs using this dataset, leaving their applicability unverified. 

MedVH \cite{gu_medvh_2024} aimed to evaluate hallucinations in medical MLLMs through False Confidence Justification (FCJ) experiments. They introduced incorrect choices into VQA prompts to assess whether MLLMs could effectively detect and correct such errors. However, FCJ presupposes the existence of clearly defined answers and is based on multiple-choice VQA that generally involves relatively low reasoning complexity.

In this study, we propose the first comprehensive Error VQA benchmark for evaluating the classification, detection, and correction capabilities of errors in radiology reports based on 3D CT images. This benchmark consists of six types of errors (omission, insertion, direction, size, unit, typo) and three difficulty levels (classification, detection, correction), providing an in-depth evaluation of the medical reasoning capabilities of currently prominent open-source 3D medical MLLMs. 
\section{Method}
\begin{table}[htbp]
\centering
\caption{The number of CT scans and question–answer triplets across different error types and task levels in the MedErr-CT benchmark.}
\label{tab:error_type_summary}
\renewcommand{\arraystretch}{1.3}
\resizebox{\columnwidth}{!}{%
\begin{tabular}{lcccc}
\toprule
\textbf{Error type} & \textbf{Classification} & \textbf{Detection} & \textbf{Correction} & \textbf{Total} \\
\midrule
Insertion     & 3,989& 3,989 &      -    &  7,978\\
Omission      & 1,902           &   -    & 1,902         & 3,804\\
Direction     &    1,942& 1,942 & 1,942        &  5,826\\
Size          &       1,390& 1,390 & 1,390         &  4,170\\
Typo          &       1,953& 1,953 & 1,953          & 5,859\\
Unit          &          1,527& 1,527 & 1,527         &  4,581\\
No Errors     & 3,000           &   3,000&       3,000& 9,000\\
\midrule
\textbf{Total}  & \textbf{15,703} & \textbf{13,801} & \textbf{11,714} & \textbf{41,218}\\
\bottomrule
\end{tabular}
}
\end{table}

\subsection{Original Dataset Preparation}
CT-RATE \cite{hamamci2024foundation} is a dataset of 25,692 non-contrast 3D chest CT volumes paired with radiology text reports derived from 21,304 patients.
RadGenome-Chest CT \cite{zhang2024radgenome} is a dataset built on CT-RATE, incorporating region-level annotations.
The MedErr-CT dataset was constructed with region-level reports from the RadGenome-Chest CT.

\begin{figure*}[t]
    \centering
    \includegraphics[width=\textwidth]{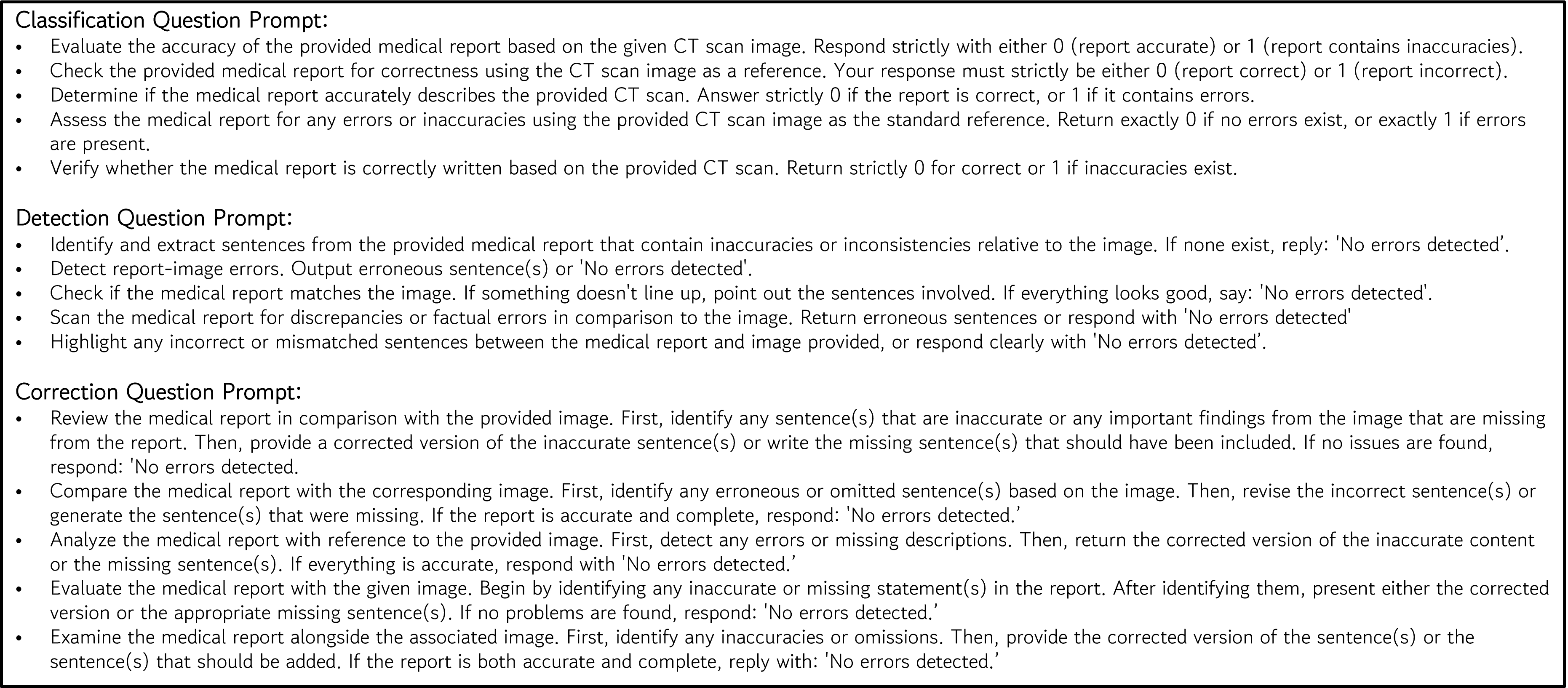}
    \caption{Question prompt list for each error task. These prompts are randomly selected during the execution of each task.}
    \label{fig:2}
\end{figure*}

\subsection{Vision-Centric Error}

Previous studies on radiological report errors primarily addressed lexical errors by arbitrarily deleting text segments or inserting irrelevant phrases into reports \cite{gertz2024potential, abacha2024medec}. However, such approaches may not consider visual context and could result in low-quality errors that are unrealistic. Lexical errors alone are insufficient to adequately assess the competency of MLLMs—an issue that has recently gained attention even in general-domain. \cite{tong2024cambrian}.

Therefore, we developed a benchmark that can accurately evaluate MLLMs' clinical reasoning capabilities by considering both lexical and vision-centric errors. To ensure high-quality error generation, we introduced errors based on findings directly observable in radiological images. We defined six categories of common and clinically significant error types: Omission, Insertion, Direction, Size, Unit, and Typo. Among these, Omission, Insertion, Direction, and Size correspond to vision-centric errors, while Unit and Typo are categorized as lexical errors.
To effectively generate error reports, we analyzed the most frequently occurring lesions in the RadGenome-Chest CT dataset. Among the most commonly observed lesions, nodules and pleural effusions are well-defined and deterministic findings, suitable for introducing meaningful errors involving the presence, location, and size of lesions. Therefore, we focused on these two clinically significant lesions.


\subsection{Benchmark Configuration and Evaluation Framework}

Inspired by the approach of MEDEC \cite{abacha2024medec}, we propose a stepwise evaluation framework that introduces various difficulty levels of errors. 
This framework consists of three levels:
\begin{itemize}
    \item \textbf{Error Classification:} Binary classification of whether an error exists in the report.
    \item \textbf{Error Detection:} Identification of the exact sentence containing the error.
    \item \textbf{Error Correction:} Correction of the erroneous sentence.
\end{itemize}

For Omission errors, error detection task could not be applied since the sentence to be detected had been removed. 
Similarly, error correction task is unnecessary for Insertion errors since detecting the errors is essentially the same as correction. For specific question-answer prompt formats, refer to Fig.~\ref{fig:2}

\subsection{Dataset Construction and Validation Process}

For dataset construction, we utilized metadata from two complementary sources describing the same patient cohort: "train\_vqa\_abnormality.csv" from the RadGenome-Chest CT dataset and "train\_metadata.csv" from the CT-RATE dataset. By combining information from both metadata files, we selected 3,000 distinct patient cases, consisting of 1,000 with nodules, 1,000 with pleural effusions, and 1,000 normal cases. For each case, error reports were generated using predefined prompts tailored to each error type. Based on these reports, error-specific QA sets were constructed using a predefined template and categorized by error levels. The final dataset comprises approximately 40,000 error-specific QA samples. The detailed distribution of the data is summarized in Table \ref{tab:error_type_summary}. To efficiently manage this error generation process, we used the DSPy framework \cite{khattab2023dspy} with LLaMA-3.3 (70B) \cite{grattafiori2024llama}. This enabled us to implement a modular LLM architecture for systematic data generation across varying error types and levels. The structure of the entire framework is illustrated in Fig.~\ref{fig:one}. Finally, to ensure the reliability of the generated errors, two radiologists manually validated the QA sets.
\newcommand{\tablefontsize}{\fontsize{10}{10}\selectfont}  

\begin{table*}[t]
\centering
\caption{Quantitative comparison of representative 3D medical MLLMs on the MedErr-CT benchmark across three evaluation levels and six error types. Each row highlights the best-performing model in red and the second-best in blue for ease of comparison. The “error” row reports the average performance across the six defined error types (omission, insertion, direction, size, unit, and typo), while the “none error” row shows the performance on undistorted, error-free reports. The “all” row reflects the overall average performance, aggregating both error and none error cases.}
\label{tab:vertical_combined_levels}
\renewcommand{\arraystretch}{1.2}
\tablefontsize
\resizebox{\textwidth}{!}{
\begin{tabular}{l|l|cccccccc}
\toprule
\textbf{Level} & \textbf{Error Type} & RadFM & M3D\textsubscript{Phi} & M3D\textsubscript{Llama}  & CT-CHAT\textsubscript{Llama} & CT-CHAT\textsubscript{Mistral} & CT-CHAT\textsubscript{Vicuna} & Med3DVLM & MedM-VL\\
\midrule 
\multirow{9}{*}{\shortstack[l]{Classification\\ \\(Accuracy)}} 
 & Insertion & 0.163& 0.000& 0.230& 0.302&\textcolor{red}{0.502}& 0.430&0.087& \textcolor{blue}{0.484}\\
 & Omission  & 0.131& 0.002& 0.252& 0.251&0.383&\textcolor{blue}{0.385}& 0.067& \textcolor{red}{0.387} \\
 & Direction & 0.136& 0.002& 0.270& 0.284&0.431& \textcolor{red}{0.461}&0.063& \textcolor{blue}{0.436}\\
 & Size      & 0.138& 0.002& 0.279& 0.302&0.445&\textcolor{blue}{0.450}& 0.058& \textcolor{red}{0.454}\\
 & Typo      & 0.137& 0.003& 0.264& 0.301&\textcolor{blue}{0.428}& \textcolor{red}{0.443}&0.065& 0.419\\
 & Unit      & 0.127& 0.004& 0.268& 0.305&\textcolor{blue}{0.403}&\textcolor{red}{0.447}& 0.077&0.393\\
\cmidrule{2-10}
 & Error     & 0.143& 0.002& 0.254& 0.292&\textcolor{red}{0.444}& 0.434&0.072& \textcolor{blue}{0.438}\\
 & None error    & 0.099& 0.397& 0.328  &\textcolor{red}{0.770}& 0.530&\textcolor{blue}{0.666}& 0.110& 0.616\\
 \cmidrule{2-10}
 & All       & 0.134& 0.077& 0.269& 0.383&0.460& \textcolor{red}{0.478}&0.080& \textcolor{blue}{0.472}\\
\midrule
\multirow{9}{*}{\shortstack[l]{Detection\\ \\(Soft Index \\Matching (SIM))}}
&Insertion  & 0.156 &0.126 & 0.095 & 0.023&\textcolor{red}{0.338} &0.076 & 0.127 & \textcolor{blue}{0.171} \\
&Direction  & \textcolor{blue}{0.203} &0.187 & 0.171 &0.009 & \textcolor{red}{0.255} &0.056 & 0.184 & 0.132 \\
&Size       & \textcolor{blue}{0.199} &0.166 & 0.139 &0.012 & \textcolor{red}{0.260} & 0.049&0.163 & 0.153 \\
&Typo       & \textcolor{red}{0.343} &\textcolor{blue}{0.337} & 0.304 &0.006 & 0.198 &0.055 & 0.283 & 0.109 \\
&Unit       & \textcolor{blue}{0.243} &0.209 & 0.199  & 0.011&\textcolor{red}{0.262}&0.058  & 0.214 & 0.148 \\
\cmidrule{2-10}
&Error       & \textcolor{blue}{0.216} &0.191   & 0.167 &0.014  & \textcolor{red}{0.277} &0.063& 0.182 & 0.147 \\
&None error     & 0.013 &0.014   & 0.108 &\textcolor{red}{0.967}  & 0.270 &\textcolor{blue}{0.829} & 0.137 & 0.639 \\
\cmidrule{2-10}
&All        & 0.172  &0.153  & 0.154 &0.220  & \textcolor{red}{0.276} &0.228  & 0.172 & \textcolor{blue}{0.253} \\
\midrule
\multirow{9}{*}{\shortstack[l]{Correction\\ \\(GREEN)}} 
 & Omission  & \textcolor{red}{0.028}&0.003& 0.000 &0.002& \textcolor{blue}{0.019}&0.003& 0.001& 0.018\\
 & Direction & \textcolor{red}{0.116}&0.014& 0.003& 0.017&\textcolor{blue}{0.110}&0.026& 0.004& 0.103\\
 & Size      & \textcolor{red}{0.130}& 0.015&0.001& 0.017&\textcolor{blue}{0.122}& 0.027&0.003& 0.118\\
 & Typo      & \textcolor{blue}{0.143}& 0.015&0.004& 0.027&\textcolor{red}{0.155}& 0.035&0.008& 0.132\\
 & Unit      & \textcolor{blue}{0.132}& 0.015&0.001& 0.024&\textcolor{red}{0.138}& 0.027&0.007& 0.129\\
\cmidrule{2-10}
 & Error     & \textcolor{red}{0.108} & 0.012&0.002& 0.017&\textcolor{blue}{0.107}& 0.024&0.005& 0.098\\
 & None error    & 0.223& 0.038&0.047& \textcolor{red}{0.884}&0.432&\textcolor{blue}{0.872}& 0.035& 0.436\\
 \cmidrule{2-10}
 & All       & 0.137& 0.019&0.014& \textcolor{blue}{0.239}&0.190& \textcolor{red}{0.241}&0.012& 0.184\\
\bottomrule
\end{tabular}}
\end{table*}
\section{Experiments}

\subsection{Evaluation details}

For evaluation, we utilized publicly available open-source 3D medical MLLM models: RadFM \cite{wu_towards_2023}, M3D \cite{bai_m3d_2024}, CT-CHAT \cite{hamamci_developing_2025}, Med3DVLM \cite{xin_med3dvlm}, and MedM-VL \cite{shi_medmvl}. We set the temperature parameter to 0 for all MLLM models to ensure consistency in text generation. We compared a total of eight model configurations, including two variants of M3D based on Phi-3-4B and Llama-2-7B language models, and four variants of CT-CHAT implementing Llama-3.1-8B, Mistral-7B, and Vicuna-7B language models. All experiments were conducted on a single NVIDIA A100 (80G) GPU, with experimental settings and hyperparameters configured according to the publicly released code.

\subsection{Zero-shot Evaluation}

We performed zero-shot evaluation for three error levels on all eight open-source MLLMs to assess their performance fairly. Questions provided with the medical images were used as instructional inputs to the MLLMs, and the generated text was compared with the reference answers. Performance for each error type and task level was comprehensively evaluated using both lexical metrics and language model-based metrics.

\subsection{Evaluation Metrics}

For the error classification task, accuracy was adopted as the primary evaluation metric. In addition, we computed precision, recall, F1-score, and specificity to provide a comprehensive assessment of model performance.

To enable fairer comparisons in error detection evaluation, we propose two novel evaluation method: Soft Index Matching (SIM) and Hard Index Matching (HIM). SIM compares model-generated output with distorted reports at the sentence level. The sentence from the distorted report that achieves the highest ROUGE-L \cite{lin_rouge_2004} score is identified as the model's answer. This matched answer is then compared with the ground-truth to determine accuracy. For HIM, we assigned an index to each sentence of the input report. The models were directly prompted to output the index corresponding to the target sentence.

To evaluate error correction performance, we employed both lexical and language model-based metrics. Lexical metrics include BLEU-4 \cite{papineni_bleu_2002}, METEOR \cite{banerjee_meteor_2005}, and ROUGE-L, which evaluate literal accuracy, fluency, and sentence structure respectively. To address limitations of these metrics in medical contexts, we incorporated two language model-based metrics: BertScore-F1 \cite{zhang_bertscore_2020}, which uses BERT embeddings to capture contextual semantic similarity, and GREEN \cite{ostmeier_green_2024}, which evaluates factual accuracy and semantic consistency on a 0-1 scale to better simulate clinical expert assessment.

\subsection{Results}
This study evaluated performance across three tasks, error classification, error detection, error correction, using an appropriate metric framework, with more detailed quantitative analyses available in \hyperref[appendix:A]{Appendix~A}, \hyperref[appendix:B]{B}, \hyperref[appendix:C]{C} and qualitative assessment in \hyperref[appendix:D]{Appendix~D}. Additionally, to evaluate the sensitivity and specificity of the models, we analyzed classification accuracy for the error and none error classes. The error class comprises six error types: Omission, Insertion, Direction, Size, Unit, and Typo, and its accuracy is calculated as the average accuracy across all six error types. The none error class consists of undistorted, error-free reports, and accuracy for this class indicates the model’s ability to correctly classify reports as error-free.

Generally speaking, our error dataset is particularly challenging due to lengthy instructions and the inherent difficulty in correcting errors based on CT images. Consequently, as shown in Table \ref{tab:vertical_combined_levels}, medical MLLMs demonstrate suboptimal performance.

According to the results of the classification task, CT-CHAT-Mistral~\cite{hamamci_developing_2025}, CT-CHAT-Vicuna, and MedM-VL~\cite{shi_medmvl} achieved the highest overall accuracies of 0.460, 0.478, and 0.472, respectively. These models also exhibited balanced performance, with recall scores of 0.444, 0.434, and 0.438, and specificity scores of 0.530, 0.666, and 0.616. In contrast, RadFM~\cite{wu_towards_2023}, M3D-Phi~\cite{bai_m3d_2024}, M3D-Llama, and Med3DVLM~\cite{xin_med3dvlm} demonstrated lower accuracies (0.134, 0.077, 0.269, and 0.080, respectively) and recall scores (0.143, 0.002, 0.254, and 0.072, respectively). The same trend was observed in F1-score. CT-CHAT-Mistral, CT-CHAT-Vicuna, and MedM-VL achieved high F1-scores of 0.571, 0.574, and 0.573, respectively, while RadFM, M3D-Phi, M3D-Llama, and Med3DVLM showed lower F1-scores of 0.210, 0.003, 0.360, and 0.113. When analyzing performance by error type, no model showed outstanding performance in any specific type compared to its overall error performance.

For error detection tasks, CT-CHAT-Mistral and MedM-VL demonstrated superior performance with overall SIM scores of 0.276 and 0.253, respectively. CT-CHAT-Mistral exhibited balanced performance across error and none error inputs (0.277 and 0.270), whereas MedM-VL showed a relatively imbalanced result (0.147 and 0.639). CT-CHAT-Llama and CT-CHAT-Vicuna achieved high scores on non-error tasks(0.967 and 0.829) while failing to detect errors(0.014 and 0.063), indicating a strong tendency to classify all inputs as error-free rather than correctly identifying erroneous cases. In contrast, RadFM, M3D-Phi, M3D-Llama, and Med3DVLM performed reasonably well on error detection (0.216, 0.191, 0.167, 0.182) but showed poor performance on none error tasks (0.013, 0.014, 0.108, 0.137), demonstrating a tendency toward over-detection. In HIM score, CT-CHAT-Llama, CT-CHAT-Mistral, CT-CHAT-Vicuna, and MedM-VL achieved reasonable performance (0.247, 0.246, 0.232, and 0.302, respectively), whereas RadFM, M3D-Phi, M3D-Llama, and Med3DVLM exhibited poor performance(0.004, 0.001, 0.002, 0.059, respectively). Notably, CT-CHAT-Mistral achieved remarkably high SIM scores in vision-centric error types compared to other models. In addition, both CT-CHAT-Mistral and MedM-VL showed relative weakness in handling typo errors compared to other error types and models.

\begin{figure}
    \centering
    \includegraphics[width=\linewidth]{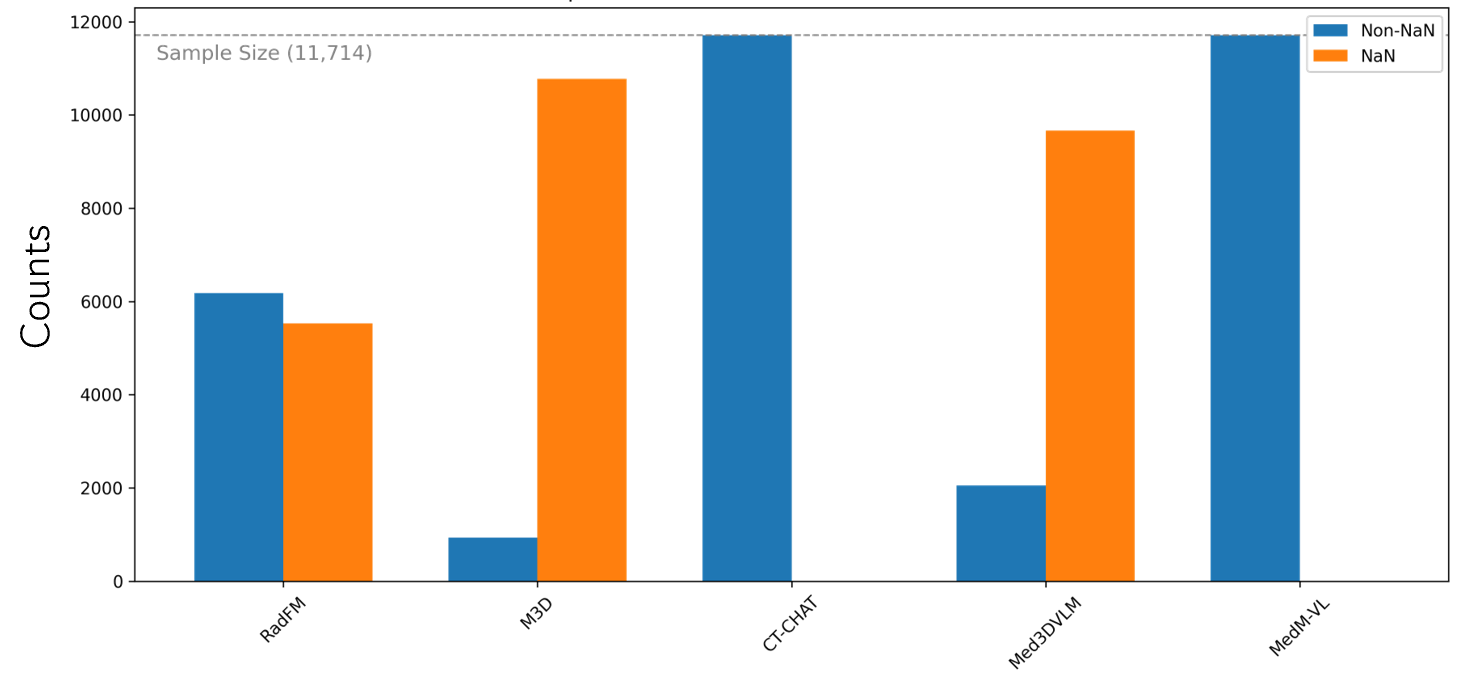}
    \caption{NaN value rates by model in correction task. (M3D = MeD-Llama, CT-CHAT = CT-CHAT-Mistral)
}
    \label{fig:3}
\end{figure}

\begin{figure}
    \centering
    \includegraphics[width=\linewidth]{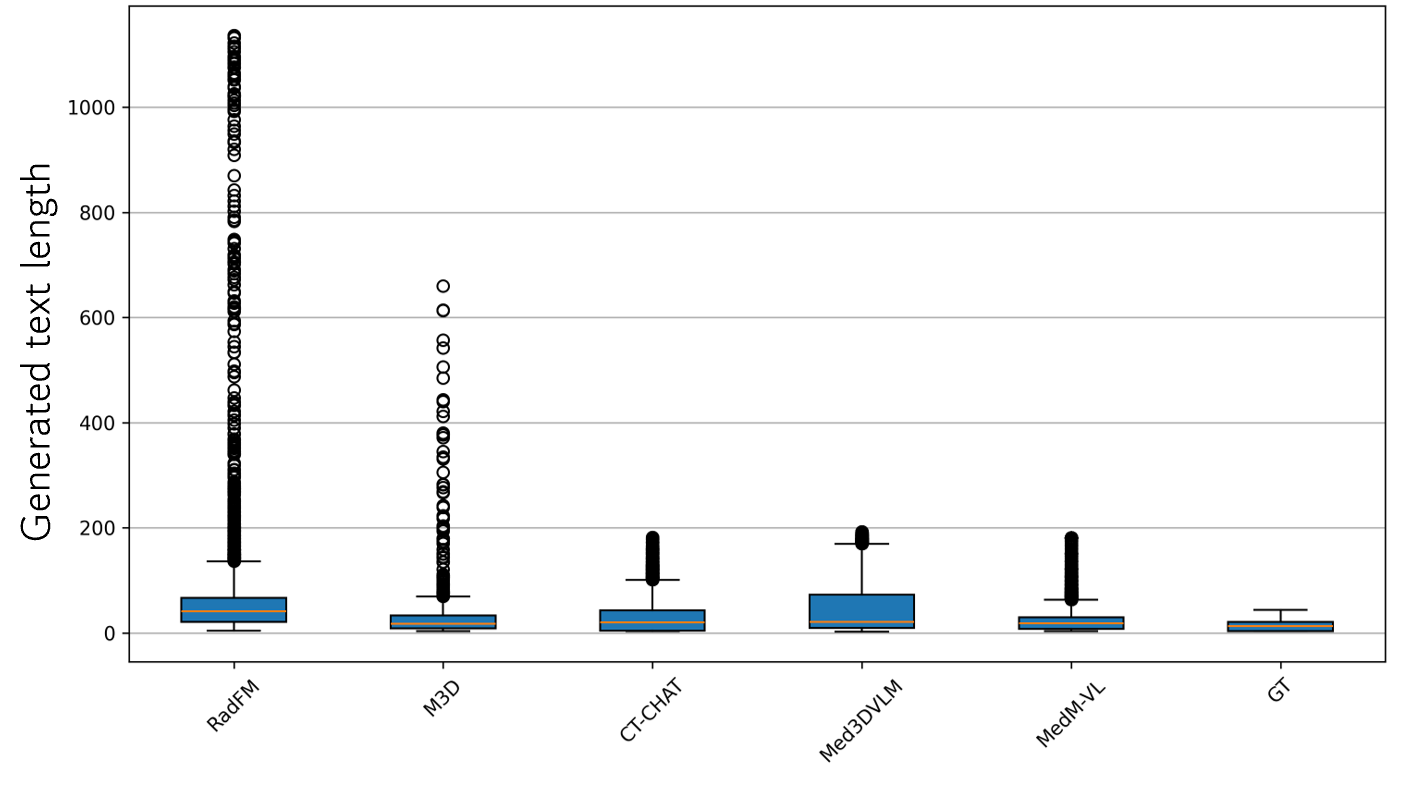}
    \caption{Generated sentence length distribution excluding NaN values. (M3D = MeD-Llama, CT-CHAT = CT-CHAT-Mistral)
}
    \label{fig:4}
\end{figure}

For the error correction task result of the GREEN metric, RadFM, CT-CHAT-Mistral, and MedM-VL achieved reasonably balanced performance for both error (0.108, 0.107, 0.098) and non-error cases (0.223, 0.432, 0.436). However, RadFM frequently produced NaN outputs and excessively long texts that did not align with the instruction as shown in Fig.~\ref{fig:3},~\ref{fig:4}, resulting in lower qualitative performance compared to CT-CHAT-Mistral and MedM-VL. Although CT-CHAT-Llama and CT-CHAT-Vicuna recorded the highest overall scores (0.239 and 0.241), the large disparity between their performance on error (0.017 and 0.024) and non-error cases (0.884 and 0.872) indicates a strong bias toward none error correction,  essentially failing to correct the error . M3D-Phi, M3D-Llama, and Med3DVLM showed poor overall correction performance (0.019, 0.014, and 0.012). In BERTScore-F1, CT-CHAT-Mistral and MedM-VL showed consistently strong performance in both error (0.852 and 0.860) and non-error cases (0.885 and 0.874), resulting in overall scores of 0.860 and 0.863, respectively. In contrast, CT-CHAT-Llama and CT-CHAT-Vicuna exhibited a large discrepancy between error performance (0.843 and 0.842) and non-error performance (0.978 and 0.971), despite achieving high overall scores of 0.877 and 0.875. RadFM, M3D-Phi, M3D-Llama, and Med3DVLM showed relatively lower overall performance (0.832, 0.820, 0.812, and 0.814). Among all error types, the omission case was the most challenging, with all models struggling to perform effectively.

\section{Discussion}

Overall, MLLMs exhibited generally suboptimal performance as the tasks became more challenging and clinically critical, with particularly poor results in the omission correction task which is especially important in clinical practice. None of the models demonstrated sufficient reliability for immediate clinical deployment. Our comparative analysis revealed that CT-CHAT and MedM-VL demonstrated superior performance compared to other models evaluated in our benchmark. We observed that the instruction-following dataset plays a more crucial role in this performance advantage than the image-text alignment dataset, training strategy, or model architecture. We also inferred that strong performance on vision-centric cases may be correlated with higher image input resolution.

\subsection{Instruction-following Dataset}

Among the compared models, CT-CHAT and MedM-VL were trained on the CT-RATE dataset, M3D and Med3D-VLM on M3D-Data, and RadFM on RadMD.

\subsubsection{Scale of Instruction-Following Dataset}

While RadMD contained the largest number of image-report pairs overall, CT-RATE included the most instruction-following data, with approximately 2.7 million data. This was followed by M3D-Data with approximately 0.67M samples and RadMD with approximately 0.55M samples in the 3D domain. The performance aligned more closely with the scale of instruction-following data rather than the total dataset size, highlighting the importance of instruction-based supervision for this task.

\subsubsection{VQA Quality of Instruction-following Dataset}
CT-RATE provided a wide range of instruction types—such as long and short answers, multiple-choice questions, and report generation—which contributed to improved model robustness.
In particular, the long answer instruction-following subset of CT-RATE, constructed as multi-turn conversational data, appears to support improved model performance.
In contrast, RadMD included multiple imaging modalities, which may have hindered performance on modality-specific tasks such as our CT-focused VQA.
M3D-Data only contained five shallow predefined question types: imaging plane, acquisition phase, organ identification, abnormality detection, and anatomical location. This simplified VQA format led to a notable performance drop on more complex tasks.
Notably, while RadMD and M3D-Data were sourced from the Radiopaedia website, CT-RATE was constructed from real-world radiology reports, providing more clinically grounded supervision.

\subsection{Image Resolution}

The CT-CHAT utilized the highest-resolution CT inputs among all models evaluated in our study, with a voxel size of 480 × 480 × 240. This likely contributed to its superior performance in vision-centric tasks compared to models using lower-resolution inputs.

\subsection{Limitations}
Despite its contributions, this study has certain limitations.

First, the current generation of MLLMs exhibits a high degree of sensitivity to prompt formulation and often fails to maintain consistency in response formats. Although diverse and systematically constructed prompts were employed, the evaluation methodology proposed in this study may not be equally optimized for all models, potentially affecting the generalizability of the results.

Second, although the question-answer pairs used in this study were newly generated and are completely independent of existing datasets, there remains a possibility of data leakage. Specifically, the visual data employed in this benchmark partially overlaps with the CT-RATE training set. Consequently, models that were trained on CT-RATE may exhibit biased performance and this potential influence cannot be entirely ruled out, even though the unseen nature is definitively confirmed.

Third, only a limited range of lesion types—such as pleural effusion and pulmonary nodules—are represented in the dataset used in this study. While the benchmark was intentionally constrained to these lesions to ensure controlled evaluation conditions, real-world clinical scenarios involve a much broader spectrum of abnormalities. Future research should therefore aim to expand the range of lesion types included in the evaluation to better reflect clinical diversity.
\section{Conclusion}

In this study, we introduce a novel benchmark to classify, detect, and correct errors in CT radiology reports. We propose a VQA framework that conducts comprehensive quantitative evaluations for MLLMs in diverse error categories of 6 error types and hierarchical error levels of 3 difficulty levels. In our evaluations of five medical MLLMs: RadFM, M3D, CT-CHAT, Med3DVLM, and MedM-VL. CT-CHAT and MedM-VL demonstrated superior performance in zero-shot tasks, underscoring the importance of diverse instructions and multi-turn conversations in model robustness.
 This study goes beyond simple hallucination evaluation to accurately assess models’ reasoning ability, making two major contributions: (1) clinically meaningful quantitative analysis of state of the art 3D medical MLLMs and (2) development of a comprehensive and clinically relevant practical error dataset. 3D medical MLLMs have not yet reached clinical-use standards, requiring benchmarks, instruction datasets, and models that handle complex clinical medical knowledge related to clinical practice. This work establishes the foundation for reliable medical MLLMs through practical insights into the error correction capabilities that clinicians genuinely require, providing direction for future model development.
{
    \small
    \bibliography{main}
    \bibliographystyle{unsrt}
}
\clearpage
\onecolumn
\appendix
\section*{Supplementary Materials}
\renewcommand{\thesection}{\Alph{section}}  
\label{sec:supplementary}

\captionsetup{justification=raggedright, singlelinecheck=false}

\renewcommand{\thetable}{A\arabic{table}}
\setcounter{table}{0}
\subsection*{\mbox{Appendix A Error Classification Result}}
\label{appendix:A}
\begin{table}[htbp]
\centering
\captionsetup{justification=raggedright, singlelinecheck=false}
\caption{Classification Metrics (of all data)}
\label{tab:level1_all}
\resizebox{\textwidth}{!}{%
\begin{tabular}{lcccccccc}
\toprule
\textbf{type} & \textbf{RadFM} & \textbf{M3D\_Phi} & \textbf{M3D\_Llama} & \textbf{CT-CHAT\_Llama} & \textbf{CT-CHAT\_Mistral} & \textbf{CT-CHAT\_Vicuna} & \textbf{Med3DVLM} & \textbf{MedM-VL} \\
\midrule
Precision        & 0.401 & 0.013 & 0.616 & 0.843 & 0.800 & \textbf{0.846} & 0.256 & 0.828 \\
Recall        & 0.143 & 0.002 & 0.254 & 0.292 & \textbf{0.444} & 0.434 & 0.072 & 0.438 \\
F1-score        & 0.210 & 0.003 & 0.360 & 0.434 & 0.571 & \textbf{0.574} & 0.113 & 0.573 \\
Specificity        & 0.099 & 0.397 & 0.328 & \textbf{0.770} & 0.530 & 0.666 & 0.110 & 0.616 \\
\bottomrule
\end{tabular}
}
\end{table}  

\captionsetup{justification=raggedright, singlelinecheck=false}

\renewcommand{\thetable}{B\arabic{table}}
\setcounter{table}{0}
\subsection*{\mbox{Appendix B: Error Detection Comparison per Error Type}}
\label{appendix:B}
\begin{table}[htbp]
\centering
\captionsetup{justification=raggedright, singlelinecheck=false}
\caption{Detection CA}
\label{tab:level2_ca}
\resizebox{\textwidth}{!}{%
\begin{tabular}{lcccccccc}
\toprule
\textbf{type} & \textbf{RadFM} & \textbf{M3D\_Phi} & \textbf{M3D\_Llama} & \textbf{CT-CHAT\_Llama} & \textbf{CT-CHAT\_Mistral} & \textbf{CT-CHAT\_Vicuna} & \textbf{Med3DVLM} & \textbf{MedM-VL} \\
\midrule
Insertion  & 0.125 & 0.024 & 0.021 & 0.007 & \textbf{0.204} & 0.048 & 0.058 & 0.199 \\
Direction  & 0.112 & 0.021 & 0.013 & 0.009 & 0.209 & 0.040 & 0.065 & \textbf{0.223} \\
Size       & 0.099 & 0.017 & 0.025 & 0.008 & \textbf{0.227} & 0.040 & 0.058 & 0.226 \\
Typo       & 0.063 & 0.008 & 0.011 & 0.002 & \textbf{0.169} & 0.036 & 0.037 & 0.146 \\
Unit       & 0.096 & 0.023 & 0.015 & 0.007 & \textbf{0.237} & 0.040 & 0.078 & 0.233 \\
\midrule
Error      & 0.104 & 0.020 & 0.018 & 0.007 & \textbf{0.206} & 0.042 & 0.058 & 0.202 \\
\bottomrule
\end{tabular}
}
\end{table}

\begin{table}[htbp]
\centering
\caption{Detection BLEU}
\label{tab:level2_bleu}
\resizebox{\textwidth}{!}{%
\begin{tabular}{lcccccccc}
\toprule
\textbf{type} & \textbf{RadFM} & \textbf{M3D\_Phi} & \textbf{M3D\_Llama} & \textbf{CT-CHAT\_Llama} & \textbf{CT-CHAT\_Mistral} & \textbf{CT-CHAT\_Vicuna} & \textbf{Med3DVLM} & \textbf{MedM-VL} \\
\midrule
Insertion & 0.022    & 0.004    & 0.006    & 0.007    & \textbf{0.081}             & 0.045    & 0.078    & 0.024    \\
Direction & 0.027 & 0.000 & 0.001 & 0.002 &0.128 & 0.017 & 0.010 &  \textbf{0.138} \\
Size      & 0.031 & 0.000 & 0.000 & 0.001 & 0.130 & 0.014 & 0.008 & \textbf{0.147} \\
Typo      & 0.022 & 0.000 & 0.001 & 0.001 & \textbf{0.111} & 0.019 & 0.011 & 0.106 \\
Unit      & 0.029 & 0.000 & 0.000 & 0.001 & 0.130 & 0.019 & 0.016 & \textbf{0.132} \\
\midrule
Error     & 0.025 & 0.001 & 0.001 & 0.002 & 0.110 & 0.025 & 0.015 & \textbf{0.112} \\
None error    & 0.002 & 0.014 & 0.013 & \textbf{0.820} & 0.088 & 0.389 & 0.011 & 0.056 \\
\midrule
All       & 0.019 & 0.002 & 0.004 & 0.022 & \textbf{0.107} & 0.057 & 0.020 & 0.103 \\
\bottomrule
\end{tabular}
}
\end{table}

\begin{table}[htbp]
\centering
\caption{Detection ROUGE-L}
\label{tab:level2_rougel}
\resizebox{\textwidth}{!}{%
\begin{tabular}{lcccccccc}
\toprule
\textbf{type} & \textbf{RadFM} & \textbf{M3D\_Phi} & \textbf{M3D\_Llama} & \textbf{CT-CHAT\_Llama} & \textbf{CT-CHAT\_Mistral} & \textbf{CT-CHAT\_Vicuna} & \textbf{Med3DVLM} & \textbf{MedM-VL} \\
\midrule
Insertion & 0.054    & 0.024    & 0.016    & 0.012    & 0.124    & 0.036    & 0.038    & \textbf{0.134}    \\
Direction & 0.046 & 0.019 & 0.011 & 0.012 & 0.128 & 0.031 & 0.035 & \textbf{0.153} \\
Size      & 0.043 & 0.015 & 0.010 & 0.006 & 0.133 & 0.028 & 0.035 & \textbf{0.147} \\
Typo      & 0.034 & 0.012 & 0.012 & 0.024 & 0.129 & 0.045 & 0.034 & \textbf{0.130} \\
Unit      & 0.040 & 0.018 & 0.009 & 0.009 & 0.135 & 0.028 & 0.042 & \textbf{0.140} \\
\midrule
Error     & 0.046 & 0.019 & 0.013 & 0.013 & 0.128 & 0.035 & 0.037 & \textbf{0.140} \\
None error    & 0.033 & 0.039 & 0.050 & \textbf{0.984} & 0.607 & 0.908 & 0.111 & 0.481 \\
\midrule
All       & 0.043 & 0.023 & 0.021 & 0.224 & \textbf{0.232} & 0.225 & 0.053 & 0.214 \\
\bottomrule
\end{tabular}
}
\end{table}

\begin{table}[htbp]
\centering
\caption{Detection BERTScore-F1}
\label{tab:level2_bertscore}
\resizebox{\textwidth}{!}{%
\begin{tabular}{lcccccccc}
\toprule
\textbf{type} & \textbf{RadFM} & \textbf{M3D\_Phi} & \textbf{M3D\_Llama} & \textbf{CT-CHAT\_Llama} & \textbf{CT-CHAT\_Mistral} & \textbf{CT-CHAT\_Vicuna} & \textbf{Med3DVLM} & \textbf{MedM-VL} \\
\midrule
Insertion & 0.833    & 0.832    & 0.829    & 0.857    & \textbf{0.863}    & 0.859    & 0.837    & 0.852    \\
Direction & 0.815 & 0.813 & 0.810 & 0.841 & \textbf{0.850} & 0.843 & 0.820 & \textbf{0.850} \\
Size      & 0.811 & 0.809 & 0.807 & 0.837 & 0.847 & 0.840 & 0.818 & \textbf{0.848} \\
Typo      & 0.808 & 0.805 & 0.805 & 0.834 & \textbf{0.843} & 0.836 & 0.814 & 0.838 \\
Unit      & 0.811 & 0.810 & 0.807 & 0.838 & \textbf{0.849} & 0.840 & 0.820 & 0.847 \\
\midrule
Error     & 0.819 & 0.818 & 0.816 & 0.845 & \textbf{0.853} & 0.847 & 0.825 & 0.848 \\
None error    & 0.827 & 0.832 & 0.834 & \textbf{0.996} & 0.927 & 0.984 & 0.842 & 0.900 \\
\midrule
All       & 0.821 & 0.821 & 0.820 & \textbf{0.878} & 0.869 & 0.877 & 0.828 & 0.859 \\
\bottomrule
\end{tabular}
}
\end{table}
\begin{table}[htbp]
\centering
\caption{Detection METEOR}
\label{tab:level2_meteor}
\resizebox{\textwidth}{!}{%
\begin{tabular}{lcccccccc}
\toprule
\textbf{type} & \textbf{RadFM} & \textbf{M3D\_Phi} & \textbf{M3D\_Llama} & \textbf{CT-CHAT\_Llama} & \textbf{CT-CHAT\_Mistral} & \textbf{CT-CHAT\_Vicuna} & \textbf{Med3DVLM} & \textbf{MedM-VL} \\
\midrule
Insertion & 0.087 & 0.030    & 0.024    & 0.055    & 0.196    & 0.086    & 0.058    & \textbf{0.220}    \\
Direction & 0.068 & 0.021 & 0.014 & 0.038 & 0.177 & 0.061 & 0.046 & \textbf{0.208} \\
Size      & 0.061 & 0.016 & 0.015 & 0.031 & 0.177 & 0.057 & 0.043 & \textbf{0.195} \\
Typo      & 0.050 & 0.014 & 0.016 & 0.047 & \textbf{0.178} & 0.073 & 0.045 & \textbf{0.178} \\
Unit      & 0.056 & 0.020 & 0.012 & 0.035 & 0.183 & 0.060 & 0.053 & \textbf{0.194} \\
\midrule
Error     & 0.069 & 0.022 & 0.018 & 0.045 & 0.185 & 0.072 & 0.051 & \textbf{0.203} \\
None error    & 0.063 & 0.049 & 0.061 & \textbf{0.969} & 0.641 & 0.912 & 0.136 & 0.552 \\
\midrule
All       & 0.068 & 0.028 & 0.027 & 0.246 & \textbf{0.284} & 0.254 & 0.069 & 0.279 \\
\bottomrule
\end{tabular}
}
\end{table}

\begin{table}[htbp]
\centering
\captionsetup{justification=raggedright, singlelinecheck=false}
\caption{Detection HIM}
\label{tab:level2_him}
\resizebox{\textwidth}{!}{%
\begin{tabular}{lcccccccc}
\toprule
\textbf{type} & \textbf{RadFM} & \textbf{M3D\_Phi} & \textbf{M3D\_Llama} & \textbf{CT-CHAT\_Llama} & \textbf{CT-CHAT\_Mistral} & \textbf{CT-CHAT\_Vicuna} & \textbf{Med3DVLM} & \textbf{MedM-VL} \\
\midrule
Insertion  & 0.006 & 0.000 & 0.001 & 0.076 & 0.116 & 0.103 & 0.004 & \textbf{0.119} \\
Direction  & 0.004 & 0.000 & 0.001 & 0.094 & 0.120 & 0.171 & 0.006 & \textbf{0.290} \\
Size       & 0.006 & 0.001 & 0.000 & 0.106 & 0.120 & 0.203 & 0.009 & \textbf{0.291} \\
Typo       & 0.006 & 0.000 & 0.001 & 0.106 & 0.143 & 0.206 & 0.013 & \textbf{0.336} \\
Unit       & 0.006 & 0.000 & 0.003 & 0.102 & 0.123 & 0.193 & 0.010 & \textbf{0.302} \\
\midrule
Error      & 0.006 & 0.000 & 0.001 & 0.092 & 0.123 & 0.160 & 0.007 & \textbf{0.237} \\
None error     & 0.000 & 0.004 & 0.005 & \textbf{0.803} & 0.689 & 0.494 & 0.243 & 0.537 \\
\midrule
All        & 0.004 & 0.001 & 0.002 & 0.247 & 0.246 & 0.232 & 0.059 & \textbf{0.302} \\
\bottomrule
\end{tabular}
}
\end{table}

\renewcommand{\thetable}{C\arabic{table}}
\setcounter{table}{0}
\FloatBarrier
\subsection*{\mbox{Appendix C: Error Correction Comparison per Error Type}}
\label{appendix:C}
\begin{table}[htbp]
\centering
\caption{Correction CA}
\label{tab:level3_ca_new_transposed}
\resizebox{\textwidth}{!}{%
\begin{tabular}{lcccccccc}
\toprule
\textbf{Metric}       & \textbf{RadFM} & \textbf{M3D\_Phi} & \textbf{M3D\_Llama} & \textbf{CT-CHAT\_Llama} & \textbf{CT-CHAT\_Mistral} & \textbf{CT-CHAT\_Vicuna} &\textbf{Med3DVLM}& \textbf{MedM-VL}  \\
\midrule
Omission  & 0.077 & 0.013 & 0.005 & 0.011 & \textbf{0.104} & 0.015  & 0.014& 0.090 \\
Direction & 0.316 & 0.049 & 0.012 & 0.046 & \textbf{0.370} & 0.089 & 0.023 & 0.315 \\
Size      & 0.326 & 0.054 & 0.008 & 0.054 & \textbf{0.392} & 0.087 & 0.023 & 0.349  \\
Typo      & 0.239 & 0.023 & 0.011 & 0.047 & \textbf{0.299} & 0.073 & 0.017 & 0.255  \\
Unit      & 0.309 & 0.051 & 0.010 & 0.059 & \textbf{0.391} & 0.072 & 0.033 & 0.333 \\
\midrule
Error     & 0.247 & 0.036 & 0.009 & 0.042 & \textbf{0.303} & 0.066 & 0.021 & 0.261 \\
\bottomrule
\end{tabular}%
}
\end{table}

\clearpage
\begin{table}[htbp]
\centering
\caption{Correction BLEU}
\label{tab:level3_bleu}
\resizebox{\textwidth}{!}{%
\begin{tabular}{lcccccccc}
\toprule
\textbf{type} & \textbf{RadFM} & \textbf{M3D\_Phi} & \textbf{M3D\_Llama} & \textbf{CT-CHAT\_Llama} & \textbf{CT-CHAT\_Mistral} & \textbf{CT-CHAT\_Vicuna} & \textbf{Med3DVLM} & \textbf{MedM-VL} \\
\midrule
Omission    & 0.013 & 0.002 & 0.001 & 0.007 & \textbf{0.027} & 0.008 & 0.002 & 0.024 \\
Direction   & 0.097 & 0.005 & 0.000 & 0.018 & 0.105 & 0.056 & 0.002 & \textbf{0.122} \\
Size        & 0.099 & 0.007 & 0.000 & 0.017 & 0.110 & 0.050 & 0.001 & \textbf{0.130} \\
Typo        & 0.075 & 0.005 & 0.000 & 0.035 & 0.114 & 0.065 & 0.006 & \textbf{0.115} \\
Unit        & 0.087 & 0.004 & 0.000 & 0.021 & 0.120 & 0.042 & 0.003 & \textbf{0.129} \\
\midrule
Error       & 0.073 & 0.005 & 0.000 & 0.020 & 0.097 & 0.046 & 0.003 & \textbf{0.105} \\
None error      & 0.007 & 0.005 & 0.005 & \textbf{0.369} & 0.030 & 0.232 & 0.002 & 0.020 \\
\midrule
All         & 0.057 & 0.009 & 0.002 & 0.052 & \textbf{0.083} & 0.080 & 0.006 & 0.082 \\
\bottomrule
\end{tabular}%
}
\end{table}

\begin{table}[htbp]
\centering
\caption{Correction ROUGE-L}
\label{tab:level3_rougel}
\resizebox{\textwidth}{!}{%
\begin{tabular}{lcccccccc}
\toprule
\textbf{type} & \textbf{RadFM} & \textbf{M3D\_Phi} & \textbf{M3D\_Llama}
              & \textbf{CT-CHAT\_Llama} & \textbf{CT-CHAT\_Mistral}
              & \textbf{CT-CHAT\_Vicuna} & \textbf{Med3DVLM} & \textbf{MedM-VL} \\
\midrule
Omission    & 0.064 & 0.026 & 0.009 & 0.022 & 0.098 & 0.024 & 0.013 & \textbf{0.114} \\
Direction   & 0.139 & 0.041 & 0.005 & 0.042 & 0.181 & 0.051 & 0.014 & \textbf{0.198} \\
Size        & 0.144 & 0.043 & 0.003 & 0.043 & 0.192 & 0.049 & 0.013 & \textbf{0.206} \\
Typo        & 0.113 & 0.034 & 0.005 & 0.060 & \textbf{0.196} & 0.061 & 0.015 & 0.194 \\
Unit        & 0.138 & 0.041 & 0.006 & 0.044 & 0.199 & 0.045 & 0.016 & \textbf{0.203} \\
\midrule
Error       & 0.117 & 0.036 & 0.006 & 0.042 & 0.171 & 0.046 & 0.014 & \textbf{0.181} \\
None error      & 0.103 & 0.025 & 0.024 & \textbf{0.881} & 0.375 & 0.843 & 0.020 & 0.281 \\
\midrule
All         & 0.114 & 0.034 & 0.010 & \textbf{0.257} & 0.223 & 0.250 & 0.016 & 0.207 \\
\bottomrule
\end{tabular}%
}
\end{table}

\begin{table}[htbp]
\centering
\caption{Correction BERTScore F1}
\label{tab:level3_bertscore}
\resizebox{\textwidth}{!}{%
\begin{tabular}{lcccccccc}
\toprule
\textbf{type} & \textbf{RadFM} & \textbf{M3D\_Phi} & \textbf{M3D\_Llama}
              & \textbf{CT-CHAT\_Llama} & \textbf{CT-CHAT\_Mistral}
              & \textbf{CT-CHAT\_Vicuna} & \textbf{Med3DVLM} & \textbf{MedM-VL} \\
\midrule
Omission    & 0.824 & 0.818 & 0.812 & 0.842 & 0.839 & 0.840 & 0.814 & \textbf{0.847} \\
Direction   & 0.835 & 0.818 & 0.806 & 0.845 & 0.856 & 0.845 & 0.810 & \textbf{0.864} \\
Size        & 0.833 & 0.815 & 0.803 & 0.842 & 0.854 & 0.842 & 0.807 & \textbf{0.864} \\
Typo        & 0.829 & 0.815 & 0.805 & 0.843 & 0.856 & 0.843 & 0.810 & \textbf{0.861} \\
Unit        & 0.832 & 0.815 & 0.804 & 0.843 & 0.856 & 0.842 & 0.808 & \textbf{0.864} \\
\midrule
Error       & 0.831 & 0.816 & 0.806 & 0.843 & 0.852 & 0.842 & 0.810 & \textbf{0.860} \\
None error      & 0.834 & 0.832 & 0.829 & \textbf{0.978} & 0.885 & 0.971 & 0.825 & 0.874 \\
\midrule
All         & 0.832 & 0.820 & 0.812 & \textbf{0.877} & 0.860 & 0.875 & 0.814 & 0.863 \\
\bottomrule
\end{tabular}%
}
\end{table}

\begin{table}[htbp]
\centering
\caption{Correction METEOR}
\label{tab:level3_meteor}
\resizebox{\textwidth}{!}{%
\begin{tabular}{lcccccccc}
\toprule
\textbf{type} & \textbf{RadFM} & \textbf{M3D\_Phi} & \textbf{M3D\_Llama}
              & \textbf{CT-CHAT\_Llama} & \textbf{CT-CHAT\_Mistral}
              & \textbf{CT-CHAT\_Vicuna} & \textbf{Med3DVLM} & \textbf{MedM-VL} \\
\midrule
Omission    & 0.103 & 0.029 & 0.011 & 0.051 & 0.143 & 0.054 & 0.018 & \textbf{0.150} \\
Direction   & 0.219 & 0.044 & 0.007 & 0.070 & \textbf{0.265} & 0.091 & 0.018 & 0.263 \\
Size        & 0.217 & 0.046 & 0.004 & 0.068 & \textbf{0.267} & 0.085 & 0.017 & 0.265 \\
Typo        & 0.179 & 0.037 & 0.008 & 0.086 & \textbf{0.282} & 0.097 & 0.022 & 0.262 \\
Unit        & 0.209 & 0.041 & 0.007 & 0.072 & \textbf{0.278} & 0.082 & 0.020 & 0.259 \\
\midrule
Error       & 0.183 & 0.039 & 0.008 & 0.069 & \textbf{0.245} & 0.082 & 0.019 & 0.238 \\
None error      & 0.170 & 0.048 & 0.036 & \textbf{0.873} & 0.449 & 0.849 & 0.035 & 0.382 \\
\midrule
All         & 0.179 & 0.041 & 0.015 & 0.275 & \textbf{0.297} & 0.278 & 0.023 & 0.275 \\
\bottomrule
\end{tabular}%
}
\end{table}

\clearpage
\captionsetup{justification=raggedright, singlelinecheck=false}

\renewcommand{\thefigure}{D\arabic{figure}}
\setcounter{figure}{0}
\subsection*{\mbox{Appendix D: Qualitative Assessment for Error Task}}
\label{appendix:D}
\vspace{-1em}  

\begin{figure}[H]
  \centering
  \includegraphics[width=\textwidth]{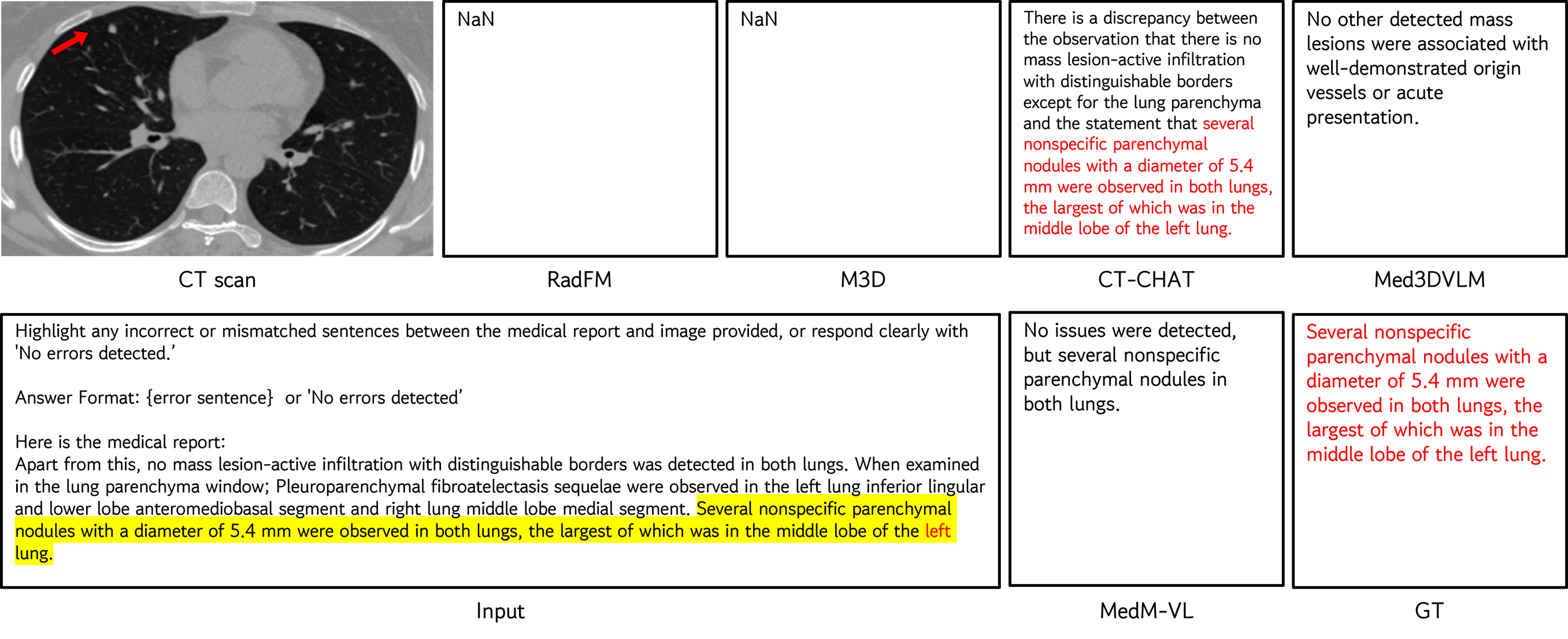}
  \caption{Qualitative assessment example of a correction task for direction error. This case involves a CT scan report of a patient with a nodule in the right lung, containing erroneous directional information. Among various models evaluated, only CT-CHAT Mistral successfully detected and corrected the erroneous statement with accuracy. (M3D = MeD-Llama, CT-CHAT = CT-CHAT-Mistral)}
  \label{fig:appendix_d_figure}
\end{figure}

\clearpage


\end{document}